\documentclass[11pt]{article}

\usepackage{alltt}
\usepackage{graphicx}
\usepackage{listings}
\usepackage{url}
\usepackage[dvipsnames]{xcolor}
\usepackage{xspace}

\usepackage[final]{automl}

\usepackage[square,numbers]{natbib}
\newcommand*{\smalltt}[1]{\texttt{\small #1}}
\newcommand*{\thetitle}[0]{A Suite of Fairness Datasets for Tabular Classification}

\hypersetup{%
  pdfauthor={Martin Hirzel and Michael Feffer}, % will be reset to "Anonymous" unless the "final" package option is given
  pdftitle={\thetitle},
  pdfsubject={},
  pdfkeywords={}
}

\makeatletter
\titleformat{\paragraph}[runin]%
            {\bfseries\normalsize}%
            {}%
            {0.5em}%
            {}%
            [\@addpunct{.}]%
\makeatother

\title{\thetitle}

\author[1]{\nameemail{Martin Hirzel}{hirzel@us.ibm.com}}
\author[2]{\nameemail{Michael Feffer}{mfeffer@andrew.cmu.edu}}
\affil[1]{IBM Research, USA}
\affil[2]{Carnegie Mellon University, Pittsburgh, PA, USA}

\begin{document}

\maketitle

\begin{abstract}
There have been many papers with algorithms for improving fairness of
machine-learning classifiers for tabular data.
Unfortunately, most use only very few datasets for their experimental
evaluation.
We introduce a suite of functions for fetching 20 fairness datasets
and providing associated fairness metadata.
Hopefully, these will lead to more rigorous experimental evaluations
in future fairness-aware machine learning research.
\end{abstract}

\begin{table}[!b]
\caption{\label{tab:static_info}Static information about the datasets.
  Column `origin' specifies from where the data is downloaded.
  Columns `\#rows' and `\#cols' give the shape of~\texttt{\small X}.
  Columns \mbox{`any cat.'} and \mbox{`any mis.'} indicate whether
  \texttt{\small X} has any categorical columns and any missing values,
  respectively.
  Column `\#labels' shows the number of unique values in \texttt{y}
  and `target name' shows the name of~\texttt{y}.
  Columns `favorable labels' and `protected attributes' are
  part of the fairness metadata.\\
  {\footnotesize *For details see \url{https://github.com/IBM/lale/blob/master/examples/demo_fairness_datasets.ipynb}}.}
\centerline{\setlength{\tabcolsep}{5pt}\scriptsize\begin{tabular}{llrrccrllll}
  \hline
  \textbf{name} & \textbf{origin} & \textbf{\#rows} & \textbf{\#cols} & \textbf{any} & \textbf{any} & \textbf{\#la-} & \textbf{target} & \textbf{favorable} & \multicolumn{2}{c}{\bf protected attributes}\\
                  &            &        &        & \textbf{cat.} & \textbf{mis.} & \textbf{bels} & \textbf{name} & \textbf{labels} & \textbf{(first)} & \textbf{(second)} \\
  \hline
  ricci           & OpenML     &    118 &      5 & yes &  no & 2 & promotion        & Promotion   &         race &            \\
  tae             & OpenML     &    151 &      5 &  no &  no & 3 & class\_attribute & 3           & whether\_...* &            \\
  heart\_disease  & OpenML     &    303 &     13 &  no &  no & 2 & target           & 1           &          age &            \\
  student\_math   & OpenML     &    395 &     32 & yes &  no & 2 & g3\_ge\_10       & 1           &          sex &         age\\
  student\_por    & OpenML     &    649 &     32 & yes &  no & 2 & g3\_ge\_10       & 1           &          sex &         age\\
  creditg         & OpenML     &  1,000 &     20 & yes &  no & 2 & class            & good        &  personal...* &         age\\
  titanic         & OpenML     &  1,309 &     13 & yes & yes & 2 & survived         & 1           &          sex &            \\
  us\_crime       & OpenML     &  1,994 &    102 & yes &  no & 2 & crimegt70pct     & 0           &  blackgt6pct &            \\
  compas\_violent & ProPublica &  4,020 &     51 & yes & yes & 2 & two\_year\_recid & 0           &          sex &        race\\
  nlsy            & OpenML     &  4,908 &     15 & yes &  no & 2 & income96gt17     & 1           &          age &      gender\\
  compas          & ProPublica &  6,172 &     51 & yes & yes & 2 & two\_year\_recid & 0           &          sex &        race\\
  speeddating     & OpenML     &  8,378 &    122 & yes & yes & 2 & match            & 1           &     samerace &  importa...*\\
  nursery         & OpenML     & 12,960 &      8 & yes &  no & 5 & class            & spec\_prior &      parents &            \\
  meps19          & AHRQ       & 16,578 &  1,825 & yes &  no & 2 & UTILIZATION      & 1           &         RACE &            \\
  meps21          & AHRQ       & 17,052 &  1,936 & yes &  no & 2 & UTILIZATION      & 1           &         RACE &            \\
  meps20          & AHRQ       & 18,849 &  1,825 & yes &  no & 2 & UTILIZATION      & 1           &         RACE &            \\
  law\_school     & OpenML     & 20,800 &     11 & yes &  no & 2 & ugpagt3          & TRUE        &        race1 &            \\
  default\_credit & OpenML     & 30,000 &     24 &  no &  no & 2 & default.pay...*  & 0           &          sex &            \\
  bank            & OpenML     & 45,211 &     16 & yes &  no & 2 & class            & 1           &          age &            \\
  adult           & OpenML     & 48,842 &     14 & yes & yes & 2 & class            & >50K        &         race &         sex\\
  \hline
\end{tabular}}
\end{table}

\section{Introduction}\label{sec:intro}

Many people share the goal of making artificial intelligence fairer
to those affected by it.
There is extensive debate about which fairness interventions are
appropriate and effective to achieve this goal.
This debate should be informed, at least in part, by rigorous
experimental evaluation.
Rigorous experiments can help stakeholders make more informed choices
among existing fairness interventions, as well as help researchers
invent better ones.
Unfortunately, most papers about fairness interventions evaluate them
on at most a handful of datasets.
This is because historically, it was hard to find and fetch datasets
relevant to fairness, as well as associate them with fairness metadata,
such as favorable labels or protected attributes.

\subsection{Related Work}

OpenML~\cite{vanschoren_et_al_2014} provides thousands of datasets
ready for machine learning experiments, but does not identify which of
them are relevant to fairness and does not provide fairness metadata.
AIF360~\cite{bellamy_et_al_2018} provides functions for fetching
8~fairness datasets along with metadata, but requires using a special
class or a multi-level pandas index.
Quy et al.~\cite{quy_et_al_2022} describe 15 fairness datasets, but do
not provide code for fetching them, do not provide machine-readable
metadata for them, and some of their datasets are difficult to obtain.
We applaud OpenML, AIF360, and Quy et al.\ for getting most of the way
towards a suite of fairness datasets and build upon their work to
take the last missing step.

\subsection{Contribution}

This paper describes a suite of 20 Python functions to fetch 20 datasets
along with fairness metadata~(see~Table~\ref{tab:static_info}).
It focuses on tabular data with classification targets, which is the
most well-studied setting.
(Other settings also have merit but are beyond the scope of this paper.)
To make these functions easy to use, they simply return data in pandas
format~\cite{mckinney_2011} along with fairness metadata in JSON format.

\section{Functions for Fetching Datasets}\label{sec:fetchers}

\begin{figure}
\begin{lstlisting}
{
   "favorable_labels": ["good"],
   "protected_attributes": [
      {
         "feature": "personal_status",
         "reference_group": ["male div/sep", "male mar/wid", "male single"]
      },
      {
         "feature": "age",
         "reference_group": [[26, 1000]]
      }
   ]
}
\end{lstlisting}
\caption{\label{fig:fairness_info}Example fairness metadata for the
  creditg dataset.
  The list \lstinline{"favorable_labels"} contains values of
  \smalltt{y} that indicate a favorable outcome.
  The list \lstinline{"protected_attributes"} contains sub-objects
  indicating the name \lstinline{"feature"} of the attribute of
  \smalltt{X}, along with a list \lstinline{"reference_group"}
  specifying which values or ranges of that attribute indicate
  membership in the privileged group.}
\end{figure}

Our suite of dataset fetching functions grew over time in an effort to
gather fairness datasets that are available for easy download with
reasonable terms and usage restrictions.
Most of them are used in the literature, and where possible, the
fairness metadata returned by our functions emulates prior work.
Each of the 20 functions does three things: first download the data,
second minimally process the data, and third provide fairness metadata
to go along with the data.

\paragraph{Download the data}
Our library distributes only functions for fetching data, but the data
itself is not part of the library.
When you consume the data, it is your responsibility to honor its
licenses and terms of use.
As shown in Table~\ref{tab:static_info}, 15 of the datasets are from
OpenML~\cite{vanschoren_et_al_2014}, which uses a \mbox{CC-BY} license as
explained here: \url{https://www.openml.org/terms}.
Our functions download these 15 OpenML datasets.
On the other hand, for the remaining 5 datasets from AHRQ and
\mbox{ProPublica}, our functions do not download the data, but instead print
instructions for downloading them manually.
Once downloaded, our functions return them from local disk.
The data use information for the MEPS datasets from AHRQ is at
\url{https://meps.ahrq.gov/data_stats/data_use.jsp}, and the \mbox{ProPublica}
data is at \url{https://github.com/propublica/compas-analysis}.
Our functions are themselves part of the open-source Lale
library~\cite{baudart_et_al_2021}, distributed under an Apache license.

\paragraph{Minimally process the data}
Our functions perform only limited preprocessing, because preprocessing
impacts fairness and can be difficult to invert.
That said, some preprocessing already happened at source before
downloading, beyond our control.
Where the prediction target is not yet categorical, our functions
discretize it.
Where necessary, our functions drop the feature column from which the
discretized target was derived.
There are some other cases where our functions drop additional feature
columns because they are not useful.
Finally, some feature columns lack a meaningful name and our functions
rename them, e.g.\ from \smalltt{"v1"} to \smalltt{"age"}.
See the code for details.

\paragraph{Provide fairness metadata}
Each of the functions returns a JSON object with fairness metadata.
Figure~\ref{fig:fairness_info} shows an example.
The fairness metadata comprises a
list of favorable labels (i.e., favorable values in \smalltt{y}) and
a list of protected attributes (i.e., column names in \smalltt{X}).
For each protected attribute, it gives a list of either ranges or
values that indicate membership in the privileged group.
In practice, features and labels relevant to
fairness considerations are subject to interpretation
and should be determined
through careful consultation with stakeholders.
Hence, we opted for a simple format that is easy to change.

\paragraph{How to use the functions}
First install the Lale library~\cite{baudart_et_al_2021} by doing
\mbox{\lstinline{pip install lale}}.
Then you can call the functions as illustrated by the following
two lines of Python code for the creditg dataset:

\begin{lstlisting}
import lale.lib.aif360
X, y, fairness_info = lale.lib.aif360.fetch_creditg_df()
\end{lstlisting}

After this code, \smalltt{X} and \smalltt{y} contain the features and
labels of the data, represented as a pandas dataframe and
series~\cite{mckinney_2011}, and
\smalltt{fairness\_info} contains the metadata, represented as a
JSON object as illustrated in Figure~\ref{fig:fairness_info}.
At this point, you can use your favorite library to split and
preprocess the data, make predictions, evaluate metrics, and perhaps
mitigate bias.
A popular choice for many of these tasks would be the sklearn
library~\cite{buitinck_et_al_2013}.
While our dataset fetching functions are part of the Lale library~\cite{baudart_et_al_2021}, you
do not need to use Lale to process their results.
On the other hand, Lale contains additional code that uses the
metadata, including bias mitigators and fairness metrics.

\begin{figure}
\centerline{\includegraphics[width=\columnwidth]{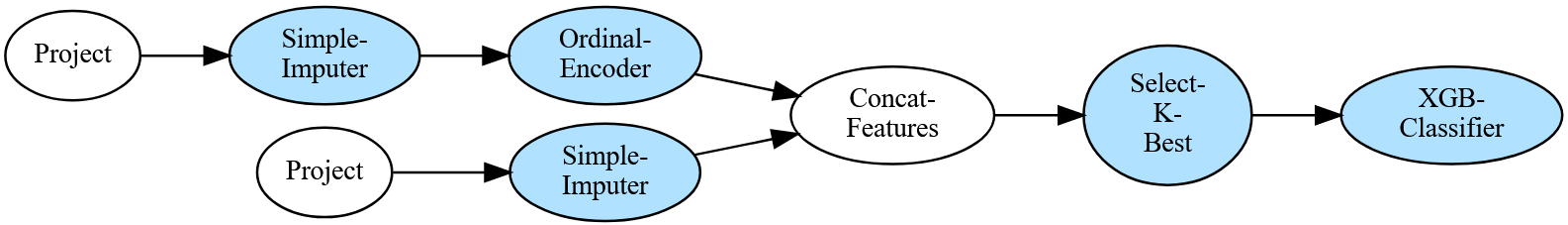}}
\caption{\label{fig:pipeline}Example pipeline for the speeddating dataset.
  (The preprocessing shown here is not part of the fetching function,
  but instead, is done on its result.)
  Referring to Table~\ref{tab:static_info},
  since \mbox{`any cat.'} is true, there are separate sub-pipelines with
  Lale \smalltt{Project} operators for categorical and numerical columns.
  Since \mbox{`any mis.'} is true, the sklearn \smalltt{SimpleImputer}
  operators fill in missing values.
  And since `\#cols' is greater than~32, an sklearn \smalltt{SelectKBest}
  operator reduces it to~32.}
\end{figure}  

\section{Characterization of the Datasets}\label{sec:results}

We already saw some static information characterizing the datasets in
Table~\ref{tab:static_info}.
This section provides additional information based on dynamic
experiments.
Our experiments feed the result from the dataset fetching function
into a pipeline of operators from Lale~\cite{baudart_et_al_2021},
sklearn~\cite{buitinck_et_al_2013}, and
\mbox{XGBoost}~\cite{chen_guestrin_2016}.
Exactly which operators are included in the pipeline depends on
whether the dataset has categorical values and missing values and
whether it has a large number of columns.
Figure~\ref{fig:pipeline} shows an example pipeline for a dataset
where all three of these are true; pipelines for other datasets are
simpler.
When you use the datasets, it is up to you what type of pipeline to use,
which may or may not resemble our example.

\begin{figure}
\centerline{\includegraphics[width=\columnwidth]{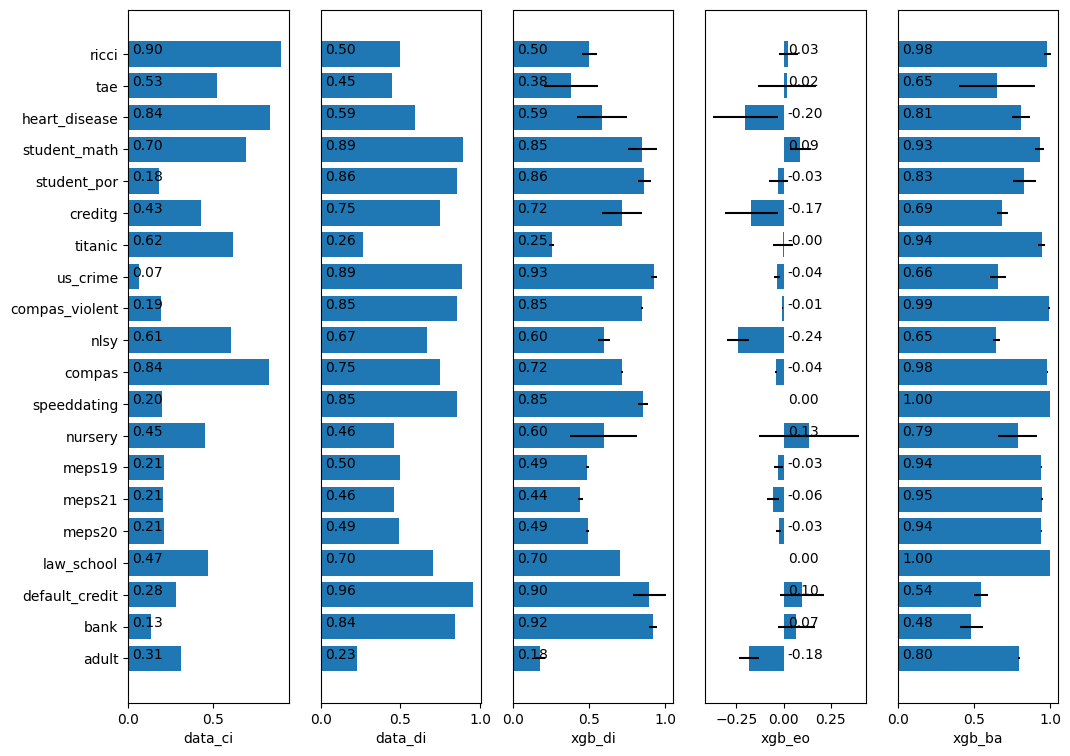}}
\caption{\label{fig:metrics}Metrics characterizing the datasets.
  Subplot `data\_ci' shows the class imbalance of the data based on
  binarizing \smalltt{y} using the metadata.
  Subplot `data\_di' shows the symmetric disparate impact of the data.
  Subplots `xgb\_di', `xgb\_eo', and `xgb\_ba' show the symmetric
  disparate impact, equal opportunity difference, and balanced
  accuracy of predictions from XGBoost.}
\end{figure}

Figure~\ref{fig:metrics} shows the results.
Subplot `data\_ci' shows class imbalance; it is the ratio between the
number of unfavorable and favorable outcomes, and higher values mean
the data is more balanced.
Subplot `data\_di' shows the symmetric disparate
impact~\cite{feldman_et_al_2015}; it is the ratio of the favorable
rates for the unprivileged and privileged groups.
Higher disparate impact values mean the data is more fair, with values
under 0.8 usually considered unfair~\cite{feldman_et_al_2015}.
The remaining three subplots show averages from 5-fold
cross-validation experiments with a popular and well-performing
classifier, XGBoost~\cite{chen_guestrin_2016}, with error bars showing
standard deviations.
Subplot `xgb\_di' shows that while bias in the data does not always
exactly equal bias in predictions of a classifier trained on the data,
the trends are similar across the 20 datasets.
Subplot `xgb\_eo' shows equal opportunity difference, which is the
difference of true positive rates between the unprivileged and
privileged groups, with zero indicating perfect fairness.
Subplot `xgb\_ba' shows balanced accuracy, which is the average recall
for the all classes, where higher values are better and the best value
is~1.
Despite using 5-fold cross-validation, the classifier overfit a couple
of datasets with 100\% balanced accuracy.

\section{Conclusion}\label{sec:conclusion}

We hope our functions for fetching fairness datasets are useful and we
welcome contributions to their open-source code.
Ideally, future papers with experimental evaluations of fairness
interventions will use at least 20, if not more, datasets.

\newpage
\bibliography{bibfile}

\end{document}